\documentclass[peerreview,onecolumn,draftcls,romanappendices]{IEEEtran}
\pdfoutput=1

\usepackage[utf8]{inputenc}
\usepackage[T1]{fontenc}
\usepackage{graphicx}
\usepackage{grffile}
\usepackage{longtable}
\usepackage{wrapfig}
\usepackage{rotating}
\usepackage[normalem]{ulem}
\usepackage{amsmath}
\usepackage{textcomp}
\usepackage{amssymb}
\usepackage{capt-of}
\usepackage{hyperref}
\usepackage{amsthm}
\usepackage{}
\hypersetup{colorlinks=true}
\renewcommand{\cal}[1]{\mathcal{#1}}
\newcommand{\op}[1]{\operatorname{#1}}

\newcommand{\Ker}[1]{\operatorname{Ker}({#1})}
\newcommand{\Ran}[1]{\operatorname{Ran}({#1})}
\newcommand{\Zer}[1]{\operatorname{Zer}({#1})}
\newcommand{\Fix}[1]{\operatorname{Fix}({#1})}
\newcommand{\diag}[1]{\operatorname{diag}({#1})}
\newcommand{\prox}[0]{\operatorname{prox}}
\newcommand{\Dom}[0]{\operatorname{Dom}}

\newtheorem{assumption}{Assumption}
\newtheorem{proposition}{Proposition}
\newtheorem{fact}{Fact}
\newtheorem{remark}{Remark}

\newtheorem{definition}{Definition}

\newtheorem{theorem}{Theorem}
\newtheorem{corollary}{Corollary}
\usepackage[margin=2cm]{geometry}
\usepackage{amssymb}
\usepackage{amsthm}
\usepackage{comment}
\usepackage{hyperref}
\usepackage{dsfont}

\begin{document}
\title{Iterative Neural Networks with Bounded Weights}
\author{Tomasz Piotrowski$^{\dagger}$ and Krzysztof Rykaczewski$^{\ddagger}$\\
  $\dagger$ Faculty of Physics, Astronomy and Informatics,\\
  Nicolaus Copernicus University,
  Grudziadzka 5/7, 87-100 Torun, Poland\\
  $\ddagger$ Faculty of Mathematics and Computer Science,\\
  Nicolaus Copernicus University,
  Chopina 12/18, 87-100 Torun, Poland\\
}

\maketitle


\begin{abstract}
A recent analysis of a model of iterative neural network in Hilbert spaces established fundamental properties of such networks, such as existence of the fixed points sets, convergence analysis, and Lipschitz continuity. Building on these results, we show that under a single mild condition on the weights of the network, one is guaranteed to obtain a neural network converging to its unique fixed point. We provide a bound on the norm of this fixed point in terms of norms of weights and biases of the network. We also show why this model of a feed-forward neural network is not able to accomodate Hopfield networks under our assumption.
\end{abstract}

\section{Introduction}
Artificial neural networks are becoming indispensible tools in a variety of spheres of human activity and society in general. It is therefore of utmost importance to understand the way they process the supplied data. We build on the recent works \cite{Combettes2019,Combettes2019a} which established fundamental properties of artificial neural networks in a general Hilbert space settings, such as existence of the fixed points sets, convergence analysis, and Lipschitz continuity. In this paper, we focus on exploiting a natural assumption that weights of the network are bounded. This assumption allows us to provide a simple proof of the fact that recurrent neural network possesses exactly one fixed cycle, obtained in the iteration limit of the network and originating from its unique fixed point. We also provide a bound on the norm of this fixed point in terms of norm of weights and biases. Finally, we discuss why a feed-forward neural network model used in this paper is not able to accomodate Hopfield networks under our assumption, which provides a motivation to derive and analyze a more generic neural network models.

\section{Preliminaries}
\subsection{Functions in Hilbert spaces}

Let $\cal{H}$ and $\cal{G}$ be arbitrary Hilbert spaces. For a function $f \colon\cal{H} \to \cal{H}$, by $\Fix{f} := \{x \mid f(x) = x\} \subset\cal{H}$ we denote fixed point set of $f$. For a set-valued function (with non-empty values) $g \colon\cal{H} \multimap \cal{G}$, $\Zer{f} := \{x \mid 0\in f(x)\}\subset \cal{H}$ stands for the set of zeros of function $f$. The closure of a subset $X \subset\cal{H}$ is denoted by $\overline{X}$. For a linear operator $A \colon\cal{H} \to\cal{G}$, by $\Ran{A} := \{Ax \mid x \in\cal{H}\} \subset\cal{G}$ and $\Ker{A} := \{x \mid Ax = 0\} \subset\cal{H}$ we denote its range and kernel, respectively. Finally, by $A^{*} \colon\cal{G} \to\cal{H}$ we denote the adjoint operator of $A$, and by $I_{\mathcal{H}} \colon \mathcal{H} \to \mathcal{H}$ we denote identity operator on $\mathcal{H}$. To simplify the notation, we omit the subscript of the identity operator if the space under consideration is clear from the context.

\subsection{Convex analysis}
Let $\mathcal{H}$ be a Hilbert space. We denote by $\Gamma_0(\cal{H})$ the class of {\em lower semi-continuous} (l.s.c.) convex functions $\phi \colon \cal{H} \to (-\infty, +\infty]$, which are {\em proper}, i.e. such that
\begin{equation}
  \Dom \phi := \{x \in \cal{H} \mid \phi(x) < +\infty\} \neq \varnothing.
\end{equation}
For $\phi \in \Gamma_0(\mathcal{H})$, the {\em proximal operator} $\prox_\phi \colon \mathcal{H} \to \mathcal{H}$ is defined as
\begin{equation}
  \op{prox}_\phi (x) := \op{argmin}_{y \in \mathcal{H}} \big( \phi(y) + 1/2 \|x - y\|^2 \big).
\end{equation}
The {\em subdifferential} of $\phi \in \Gamma_0(\mathcal{H})$ is the set-valued operator $\partial \phi \colon \cal{H} \multimap \cal{H}$ given by
\begin{equation}
  \partial \phi(x) := \{u \in \mathcal{H} \mid \langle y - x, u \rangle + \phi(x) \leq \phi(y) \enspace \mbox{for all} \enspace y \in \mathcal{H}\}.
\end{equation}
The sets $\partial \phi(x)$ are closed and convex \cite[Proposition 16.4]{Bauschke2017}.

Operator $A \colon \mathcal{H} \multimap\mathcal{H}$ is called {\em monotone}, provided for each $(x, u), (x', u') \in \op{gra}(A) := \{(y, v) \mid v \in A y\}$ we have $\langle x - x', u - u' \rangle \geq 0$. Monotone operator $A$ is {\em maximally monotone} if there is no extension of $A$ to larger monotone operator $B \colon \mathcal{H} \multimap\mathcal{H}$ in the sense that $\op{gra}(A) \subsetneq \op{gra}(B)$. An example of maximally monotone operator is the subdifferential $\partial \phi.$

Denote by $\cal{A}(\mathbb{R})$ the set of functions from $\mathbb{R}$ to $\mathbb{R}$, which are increasing, $1$-Lipschitz, and take value $0$ for argument $0$. This set of functions can be characterized using proximal operators as follows:
\begin{proposition} \cite[Proposition 2.3]{Combettes2019}
  Let $\rho \colon \mathbb{R} \to \mathbb{R}$. Then $\rho \in \cal{A}(\mathbb{R})$ if and only if there exists $\phi \in \Gamma_0(\mathbb{R})$, which has $0$ as its minimizer and $\rho = \prox \phi$.
\end{proposition}

\noindent This fact allows the following definition:

\begin{definition}\cite[Definition 2.20]{Combettes2019} \label{cals}
  Let $\cal{H}$ be real Hilbert space and let $\sigma \colon \cal{H} \to \cal{H}$. Then $\sigma$ belongs to $\cal{A}(\cal{H})$ if there exists a function $\phi \in \Gamma_0(\cal{H})$ such that it has minimium at $0$ and $\sigma = \prox \phi$.
\end{definition}

\section{Setting}

Let $n \in \mathbb{N}$, and let $(\cal{H}_i)_{0 \leq i \leq  n}$ be real Hilbert spaces. 
Let $\mathcal{H} := \cal{H}_1 \otimes \cdots \otimes \cal{H}_{n - 1} \otimes \cal{H}_n$, i.e. if $x \in \mathcal{H}$, then $x = (x_1, \ldots, x_n)$, $x_i \in \mathcal{H}_i$, $i = 1, \ldots, n$, and if $y \in \mathcal{H}$, then $\langle x, y \rangle_{\mathcal{H}} := \sum_{i = 1}^n \langle x_i, y_i \rangle_{\mathcal{H}_i}$. For simplicity we will write $\langle \cdot, \cdot \rangle$ instead of $\langle x, y \rangle_{\mathcal{H}}$.

\begin{assumption} \label{assum1}
  Let $\sigma_i \in \cal{A}(\cal{H}_i)$, i.e. $\sigma_i = \prox \phi_i$, for a certain $\phi_i \in \Gamma_0(\cal{H}_i)$ with $\phi_i(0) = \inf_{x \in \cal{H}_i} \phi_i(x)$. 
\end{assumption}

\begin{assumption} \label{assum2}
Let $W_i \colon \mathcal{H}_{i - 1} \to \mathcal{H}_i$, $i = 1, \ldots, n$ be bounded linear operators and $b_i \in \mathcal{H}_i$, $i = 1, \ldots, n$. Moreover, let us define $g_{i} \colon \mathcal{H}_{i - 1} \to \mathcal{H}_i$ by the formula
\begin{equation}
  g_{i}(x) := \sigma_{i} (W_{i} x + b_{i}), \quad \mbox{for} \enspace x \in \mathcal{H}_{i - 1}, i = 1, \ldots, n.
\end{equation}
\end{assumption}

\label{sec:org206d06a}
\begin{definition}
  An {\em $n$-layer feed-forward neural network} defined on $(\mathcal{H}_i)_{0 \leq i \leq n}$ is the composition
  \begin{equation}\label{eq:nn}
    g_n \circ \cdots \circ g_1.
  \end{equation}
\end{definition}

\noindent In the theory of neural networks, the functions $\sigma_i$ are called {\em activation operators}, operators $W_i$ are called {\em weight operators} and elements $b_i$ are called {\em bias parameters}.

\begin{remark}
 We note that Assumption \ref{assum2} implies that the neural network is already trained, by which we understand that there exists an optimal setting of weight operators $W_i$ and bias parameters $b_i$ which fits the input and output of the network to a certain set of training data. In the case of recurrent neural networks, we have a situation where the input and output of the network are in the same space $\mathcal{H}_0$, so it is natural to ask about the existence of periodic points and about the shape of set of these points.
\end{remark}
  
Thus, from now on we assume that $\cal{H}_n = \cal{H}_0$ and denote
\begin{equation}
  \mathcal{G} := \Fix{g_n \circ \cdots \circ g_1} \subset \cal{H}_0,
\end{equation}
and
\begin{equation} \label{F}
    \mathcal{F} := \{ (x_1, \ldots, x_n) \in \mathcal{H} \mid x_1 = g_1(x_n), x_2 = g_2(x_1), \ldots, x_n = g_n (x_{n - 1})\}.
\end{equation}
The set $\mathcal{G} = \Fix{g}$ consists of fixed (periodic) points of the recurrent neural network
\begin{equation} \label{g}
  g := g_n \circ \cdots \circ g_1 \colon \mathcal{H}_0 \to \mathcal{H}_1 \to \cdots \to \mathcal{H}_n = \mathcal{H}_0,
\end{equation}
and the set $\mathcal{F}$ describes trajectories across layers of the neural network \eqref{eq:nn} of these fixed points.

Let $\vec{\mathcal{H}} := \cal{H}_n \otimes \cal{H}_1 \otimes \cdots \otimes \cal{H}_{n - 1} $ and let us introduce operators
\begin{align*}
  S & \colon \mathcal{H} \to \vec{\mathcal{H}} \colon (x_1, \ldots, x_{n - 1}, x_n) \mapsto \big(x_n, x_1, \ldots, x_{n - 1}\big), \\
  W & \colon \vec{\mathcal{H}} \to \mathcal{H} \colon (x_n, x_1, \ldots, x_{n - 1}) \mapsto \big(W_1 x_n, W_2 x_1, \ldots, W_n x_{n - 1}\big).
\end{align*}
Observe that $\|W \circ S\| = \|W\| := \max_{i = 1, \ldots, n} \|W_i\|$.

Let $\phi_i\in\Gamma_0({H}_i)$ for $i = 1, \ldots,n$ 
and let us define $\phi := \phi_1 \oplus \cdots \oplus \phi_n \colon \mathcal{H} \to (-\infty, +\infty]$ by the formula
\begin{equation}
    \phi (x_1, \ldots, x_n) := \sum_{i = 1}^{n} \phi_i(x_i), \quad x_i \in \mathcal{H}_i.
\end{equation}
Moreover, let $\psi \colon \mathcal{H} \to (-\infty, +\infty]$ be defined by the formula
\begin{equation}
    \psi (x_1, \ldots, x_n) := \sum_{i = 1}^{n} \big(\phi_i(x_i) - \langle x_i, b_i \rangle \big) = \phi(x) - \langle x, b \rangle,
\end{equation}
where we denoted $x := (x_1, \ldots, x_n)$, $b := (b_1, \ldots, b_n) \in \mathcal{H}$. We also note that $\phi,\psi\in\Gamma_0(\cal{H}).$

\begin{fact}\cite[Proposition 16.9]{Bauschke2017}
  Under the above assumptions one has
  \begin{equation}
    \partial \phi(x_1, \ldots, x_n) = \partial \phi_1(x_1) \times \ldots \times \partial \phi_n(x_n).
  \end{equation}
\end{fact}
\noindent Consequently, $\partial \psi(x) = \partial \phi(x) - b$.

\begin{theorem}\cite[Part of Proposition 4.3]{Combettes2019} \label{ThC}
In terms of the model introduced above, consider the following problem: find 
 $\overline{x}_1 \in \cal{H}_1, \ldots, \overline{x}_n \in \cal{H}_n$ such that
\begin{equation} 
    \begin{cases}
    b_1 \in \overline{x}_1 - W_1 \overline{x}_n + \partial \phi_1(\overline{x}_1), \\
    b_2 \in \overline{x}_2 - W_2 \overline{x}_1 + \partial \phi_2(\overline{x}_2), \\
    \vdots \\
    b_n \in \overline{x}_n - W_n \overline{x}_{n - 1} + \partial \phi_n(\overline{x}_n).
    \end{cases}
    \label{eq:sol}
\end{equation}
The following holds.
\begin{enumerate}
    \item The set of solutions of system of inclusions \eqref{eq:sol} is $\mathcal{F}$.
    \item $\mathcal{F} = \Zer{I - W \circ S + \partial \psi} = \Fix{\prox_{\psi} \circ W \circ S}$.
    \item Let us assume that the operator $I - W \circ S$ is monotone. Then the set $\mathcal{F}$ is closed and convex. Moreover, $\mathcal{G}$ and $\mathcal{F}$ are nonempty if any of the following conditions is satisfied:
    \begin{enumerate}
        \item $I - W \circ S + \partial\phi$ is surjective.
        \item $\|W\|\leq 1$, $\Ran{S^{*} - W} = \overline{\Ran{S^{*} - W}}$ and $\Ker{S - W^{*}} = \{0\}$.
    \end{enumerate}
\end{enumerate}
\end{theorem}

\begin{remark}
  Note that problem of finding fixed points of the recurrent neural network \eqref{eq:nn} reduces to a problem of solving a system of equations
  \begin{equation}
      \begin{cases}
      \overline{x}_1 = g_1(\overline{x}_n) = \sigma_1 (W_1\overline{x}_n + b_1), \\
      \overline{x}_2 = g_2(\overline{x}_1) = \sigma_2 (W_2\overline{x}_1 + b_2), \\
      \vdots \\
      \overline{x}_n = g_n(\overline{x}_{n-1}) = \sigma_n (W_n\overline{x}_{n-1} + b_n). \\
      \end{cases}
      \label{eq:sos}
  \end{equation}
  Under Assumption \ref{assum1} and using \cite[Proposition 16.44]{Bauschke2017}, which states that
  \begin{equation}
    \prox \phi_i = (I + \partial \phi_i)^{-1},
  \end{equation}
  the above system \eqref{eq:sos} can be rewritten to the system of inclusions \eqref{eq:sol}. That is why that inclusion is crucial for our further considerations.
\end{remark}

\section{Results}
The following proposition shows that, under a single mild assumption, the neural network $g$ converges to its unique fixed point.
\begin{proposition} \label{p1}
Let 
\begin{equation} \label{W}
  \prod_{i = 1}^{n} \|W_i\| < 1.
\end{equation}
  Then, $\cal{F}$ is a singleton. Denote the unique element of $\cal{F}$ as $x^{\mathcal{F}}$. Furthermore, let $x_0 \in \cal{H}_0$ and $x_k = g^k(x_0).$ Then $\lim_{k \to \infty}x_k = x^{\mathcal{F}}_{n}$, where $x^{\mathcal{F}}_{n} \in \mathcal{H}_n$ denotes the $n$th coordinate of $x^{\mathcal{F}}$, and $\cal{G} = \{x^{\mathcal{F}}_{n}\}.$ 
\end{proposition}
\proof By Assumption \ref{assum1}, from \cite[Proposition 12.28]{Bauschke2017}, the activation operators are firmly nonexpansive. Thus, from \cite[Proposition 3.3]{Combettes2019a}, we obtain in particular that the $n$-layered neural network $g$ in \eqref{g} is Lipschitz continuous with constant $\theta_n := \prod_{i = 1}^{n} \|W_i\| < 1.$ Therefore, from the Banach Fixed Point Theorem it admits a unique fixed point, thus $\cal{G}$ is a singleton. Denote this unique fixed point of $g$ as $x^{\mathcal{F}}_{n} := \lim_{k \to \infty} x_k.$ Then, the fact that $\cal{F}$ is a singleton follows immediately from the definition of $\cal{F}$ in \eqref{F}. We also note \emph{en passant} that $\cal{F} = \big\{\big(g_1(x^{\mathcal{F}}_{n}), g_2(g_1(x^{\mathcal{F}}_{n})), \ldots, x^{\mathcal{F}}_{n}\big)\big\}.$ \qed

\begin{remark}
The above proposition extends the results of Theorem \ref{ThC} proved in \cite{Combettes2019} to the case when condition \eqref{W} holds. It is remarkable that, in such a case, no other conditions are required to ensure that $\cal{F}$ is not only nonempty, closed and convex, but actually a singleton. Moreover, note that the sequence of interations convergences in the strong topology to the unique fixed point. 
\end{remark}  

\begin{corollary}
  In particular, if
  \begin{equation} \label{Wbis}
    \|W\| < 1,
  \end{equation}
  then $\cal{F}$ is a singleton.
\end{corollary}

The next proposition provides a bound on the norm of the unique element of $\cal{F}.$
\begin{proposition} \label{p2}
Let $W$ satisfies condition \eqref{Wbis}. Then, the unique element $x^{\mathcal{F}} \in \mathcal{F}$ (\emph{cf.} Proposition \ref{p1}) is such that
\begin{equation} \label{am2}
  \|x^{\mathcal{F}}\| \leq\frac{\|b\|}{1-\|W\|}.
\end{equation}
In particular, if no bias terms are used in neural network \eqref{eq:nn}, i.e., $b = 0$, then $\mathcal{F} = \{0\}.$
\end{proposition}
\proof We note first that $x^{\mathcal{F}} = (x^{\mathcal{F}}_1, \ldots, x^{\mathcal{F}}_{n - 1}, x^{\mathcal{F}}_n)$ is the unique solution of system \eqref{eq:sol}, if condition \eqref{W} is assumed, and in such a case
\begin{equation}
  b + (W \circ S) x^{\mathcal{F}} \in x^{\mathcal{F}} + \partial \phi(x^{\mathcal{F}}) = (I + \partial\phi)(x^{\mathcal{F}}).
\end{equation}
Thus, from \cite[Proposition 16.44]{Bauschke2017} one has
\begin{equation}
  x^{\mathcal{F}} = \prox_{\phi}(W \circ S x^{\mathcal{F}} + b).
\end{equation}
Hence, using the fact that $\prox_{\phi}\in\cal{A}(\mathcal{H})$ (\emph{cf.} Definition \ref{cals}), from \cite[Proposition 2.21]{Combettes2019} one has in particular that for all $x \in \mathcal{H}$
\begin{equation}
  \|\prox_{\phi}(x)\| \leq \|x\|.
\end{equation}
For $x^{\mathcal{F}}$, this implies that
\begin{align}
  \|x^{\mathcal{F}}\| \leq & \|W \circ S x^{\mathcal{F}} + b\|\leq \|W \circ S x^{\mathcal{F}}\| + \|b\| \leq \\
  & \|W \circ S\| \cdot \|x^{\mathcal{F}}\| + \|b\| = \|W\| \cdot \|x^{\mathcal{F}}\| + \|b\|,
\end{align}
where we have used the fact that $\|W\circ S\|=\|W\|.$ The inequality \eqref{am2} follows. If $b = 0$, then $x^{\mathcal{F}} = 0$, and from Proposition \ref{p1} one concludes that in such a case $\mathcal{F} = \{0\}.$ \qed

The fact that neural network $g$ converges to a single fixed point may not be desirable in certain applications. The following remark demonstrates that in such a case, a more general network model must be considered.
\begin{remark}
  Consider Hopfield neural network model given as follows
  \begin{equation} \label{eq:hopfield}
    x'(t) = -D x(t) + W \sigma\big(x(t)\big) + b,
  \end{equation}
  where $x(t) \in \mathcal{H}$ is the state vector, $D := \diag{d_1 I_{\mathcal{H}_1}, d_2 I_{\mathcal{H}_2}, \ldots, d_n I_{\mathcal{H}_n}}$ with $d_i > 0$ is block diagonal matrix of self-inhibition of neurons and $\sigma$, given by $\sigma(x_1, \ldots, x_n) := \big( \sigma_1(x_1), \ldots, \sigma_n(x_n) \big)$, $x_i \in \mathcal{H}_i$, $i = 1, \ldots, n$, is continuous activation function of the neural network.

  Any equlibrium point $x$ of the above network satisfies
  \begin{equation}
    0 = -D x + W \sigma(x) + b.
  \end{equation}
  Therefore,
  \begin{equation}
    x = D^{-1}\big(W \sigma(x) + b\big).
  \end{equation}
  Denote $z = \sigma(x)$. Since, by \cite[Proposition 16.44]{Bauschke2017}, $x \in \sigma^{-1}(z) = z + \partial \phi(z)$, then
  \begin{equation}
    D^{-1}\big(W z + b\big) \in z + \partial \phi(z).
  \end{equation}
  Thus,
  \begin{equation}
    D^{-1}b \in z - D^{-1}W z  + \partial \phi(z),
  \end{equation}
  which is of the form \eqref{eq:sol}. In particular, under appropriate assumptions on operators $D$ and $W$, one can achieve that $\|D^{-1}W\|<1$, which according to Proposition \ref{p1} leads to a situation where we have only one fixed point of network \eqref{eq:hopfield}. Hence, our (and the one described in \cite{Combettes2019}) model may not be adequate in this case, as Hopfield network learning relies on memorizing many distinct fixed points of the network.
\end{remark}


\begin{thebibliography}{9}
  \bibitem{Combettes2019} P. L. Combettes and J.-C. Pesquet, \textit{Deep Neural Network Structures Solving Variational Inequalities}, arXiv preprint, arXiv:1808.07526, 2019.
  \bibitem{Bauschke2017} H. H. Bauschke and P. L. Combettes, \textit{Convex Analysis and Monotone Operator Theory in Hilbert Spaces}, Second Edition. New York: Springer, 2017.
  \bibitem{Combettes2019a} P. L. Combettes and J.-C. Pesquet, \textit{Lipschitz Certificates for Neural Network Structures Driven by Averaged Activation Operators}. arXiv preprint arXiv:1903.01014, 2019.
\end{thebibliography}
\end{document}